\documentclass[nohyperref]{article}

\usepackage{microtype}
\usepackage{graphicx}
\usepackage{subfigure}
\usepackage{booktabs} 

\usepackage{hyperref}

\usepackage[accepted]{icml2022}
\usepackage{microtype}
\usepackage{graphicx}
\usepackage{subfigure} 
\usepackage{multirow}
\usepackage{threeparttable}
\usepackage{microtype}
\usepackage{tabulary}
\usepackage{array}
\usepackage{pifont}
\usepackage{tabularx}
\usepackage{adjustbox}
\usepackage{multirow}
\usepackage{enumitem}
 \usepackage{xspace}
\usepackage{tcolorbox}
\usepackage{booktabs,amsfonts,dcolumn}
\usepackage{hyperref}
\usepackage{url}
\usepackage{amsmath,amsthm,amsfonts,amssymb,bm,stmaryrd,bbm}

\usepackage{makecell}
\usepackage{amsmath}
\usepackage{amssymb}
\usepackage{mathtools}
\usepackage{amsthm}

\usepackage[capitalize,noabbrev]{cleveref}

\theoremstyle{plain}

\theoremstyle{definition}

\theoremstyle{remark}

\newcommand\name{TLM}
\newcommand\ours{\name\xspace}

\newcommand\bertcorpus{$\mathcal{C}_{\text{BERT}}$\xspace}
\newcommand\robertacorpus{$\mathcal{C}_{\text{RoBERTa}}$\xspace}

\newcommand\hl[1]{{\color{blue}\textit{#1}}}

\newcommand\tf[1]{\textbf{#1}}

\newcommand{\std}{\small$\pm$}
\newcommand\eir{$\rho_1$\xspace}
\newcommand\mw{$\rho_2$\xspace}

\usepackage[textsize=tiny]{todonotes}

\icmltitlerunning{NLP From Scratch Without Large-Scale Pretraining: A Simple and Efficient Framework}

\begin{document}

\twocolumn[
\icmltitle{NLP From Scratch Without Large-Scale Pretraining: \\ A Simple and Efficient Framework}



\icmlsetsymbol{equal}{*}

\begin{icmlauthorlist}
\icmlauthor{Xingcheng Yao}{equal,iiis}
\icmlauthor{Yanan Zheng}{equal,cst}
\icmlauthor{Xiaocong Yang}{sch,rai}
\icmlauthor{Zhilin Yang}{iiis,sqz,rai}
\end{icmlauthorlist}

\icmlaffiliation{iiis}{Institute for Interdisciplinary Information Sciences, Tsinghua University}
\icmlaffiliation{cst}{Department of Computer Science and Technology, Tsinghua University}
\icmlaffiliation{sch}{School of Economics and Management, Tsinghua University}
\icmlaffiliation{sqz}{Shanghai Qi Zhi Institute}
\icmlaffiliation{rai}{Recurrent AI, Inc}

\icmlcorrespondingauthor{Zhilin Yang}{zhiliny@tsinghua.edu.cn}

\icmlkeywords{Pretraining, Self-Supervised Learning}

\vskip 0.3in
]



\printAffiliationsAndNotice{\icmlEqualContribution} 

\begin{abstract}
Pretrained language models have become the standard approach for many NLP tasks due to strong performance, but they are very expensive to train. We propose a simple and efficient learning framework \ours that does not rely on large-scale pretraining\footnote{In the broadest sense, pretraining means training on some objectives before optimizing the target tasks. In contrast, throughout the paper, we use ``pretraining'' to only refer to task-agnostic training of language models on a large general corpus, such as BERT \cite{devlin2018bert}.}. Given some labeled task data and a large general corpus, \ours uses task data as queries to retrieve a tiny subset of the general corpus and jointly optimizes the task objective and the language modeling objective from scratch. On eight classification datasets in four domains, \ours achieves results better than or similar to pretrained language models (e.g., RoBERTa-Large) while reducing the training FLOPs by two orders of magnitude. With high accuracy and efficiency, we hope \ours will contribute to democratizing NLP and expediting its development \footnote{Our code,  model checkpoints and datasets are publicly available at: \url{https://github.com/yaoxingcheng/TLM}}.
\end{abstract}


\section{Introduction}

Pretrained language models (PLMs) have drawn much attention from the natural language processing (NLP) community. Neural networks based on the Transformer architecture~\cite{vaswani2017attention} are trained on large general corpora for self-supervised language modeling tasks such as masked language modeling~\cite{devlin2018bert,liu2019roberta,raffel2019exploring}, autoregressive language modeling~\cite{radford2018gpt,brown2020language}, permutation language modeling~\cite{yang2020xlnet}, etc, and then are finetuned on a small amount of labeled data for downstream tasks. This pretraining-finetuning framework has significantly improved the performance of many NLP tasks.

However, while considered effective, large-scale pretraining is usually computationally expensive. For example, RoBERTa-Large~\cite{liu2019roberta}, a widely-used PLM, consumes a computational cost of $4.36\times10^{21}$ 
FLOPs\footnote{It was pretrained with 1,000 V100 GPUs each with 32GB memory for approximately one day.}.
Larger PLMs such as GPT-3~\cite{brown2020language} consume 50 times more FLOPs for training than RoBERTa-Large. 
The expensiveness of large-scale pretraining prevents many research groups with limited budgets from pretraining customized language models, exploring new neural architectures, or improving pretraining loss functions. In contrast, a large number of NLP researchers resort to improving the finetuning algorithms, whose performance is largely upper-bounded by the pretraining procedure. This creates a high barrier of NLP research and might not be ideal for the long-term development of the field.

\begin{figure}[t]
    \centering
    \includegraphics[width=\columnwidth]{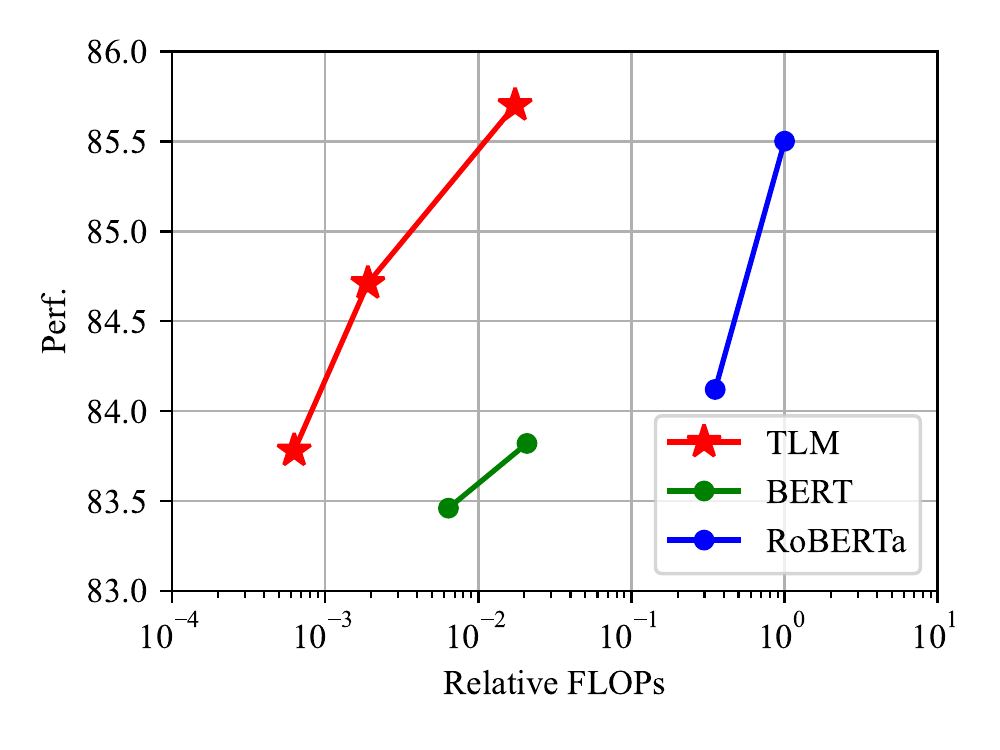}
    \vspace{-30pt}
    \caption{
    Average performance on eight tasks v.s. relative FLOPs w.r.t. RoBERTa-Large \cite{liu2019roberta}. \ours slightly outperforms RoBERTa-Large while reducing FLOPs by two orders of magnitude.}
    \label{fig:overview_result}
    \vspace{-10pt}
\end{figure}

Even though there have been efforts devoted to studying and improving the efficiency of language model pretraining~\cite{clark2020electra,so2021primer,tay2021scale,Chen_2021}, most of them focus on designing sample-efficient self-supervised tasks or discovering efficient Transformer architectures suitable for pretraining.
Their improvements are limited, with a reduction of computational costs (in terms of FLOPs) less than one order of magnitude.
Another line of works target reducing the sizes of PLMs using distillation \cite{sanh2019distilbert,jiao2019tinybert} to improve the efficiency of inference, but these methods rely on pretraining a large PLM before distillation. Moreover, distilled models often do not perform as well as some of the best non-distilled PLMs such as RoBERTa-Large \cite{sanh2019distilbert,jiao2019tinybert}.

This work explores alternatives to the standard pretraining-finetuning paradigm, aiming at more drastic efficiency improvement without performance drop. We propose a simple, efficient, pretraining-free framework, \textbf{T}ask-driven \textbf{L}anguage \textbf{M}odeling (\textbf{\ours}). Given a large general corpus and some labeled task data, \ours directly trains a model from scratch without relying on PLMs. \ours is motivated by two key ideas. First, humans master a task by using only a small portion of world knowledge (e.g., students only need to review a few chapters, among all books in the world, to cram for an exam). We hypothesize that there is much redundancy in the large corpus for a specific task. Second, training on supervised labeled data is much more data efficient for downstream performance than optimizing the language modeling objective on unlabeled data.
Based on these motivations, \ours uses the task data as queries to retrieve a tiny subset of the general corpus. This is followed by jointly optimizing a supervised task objective and a language modeling objective using both the retrieved data and the task data.

\begin{figure*}[ht]
    \centering
    \includegraphics[width=\textwidth]{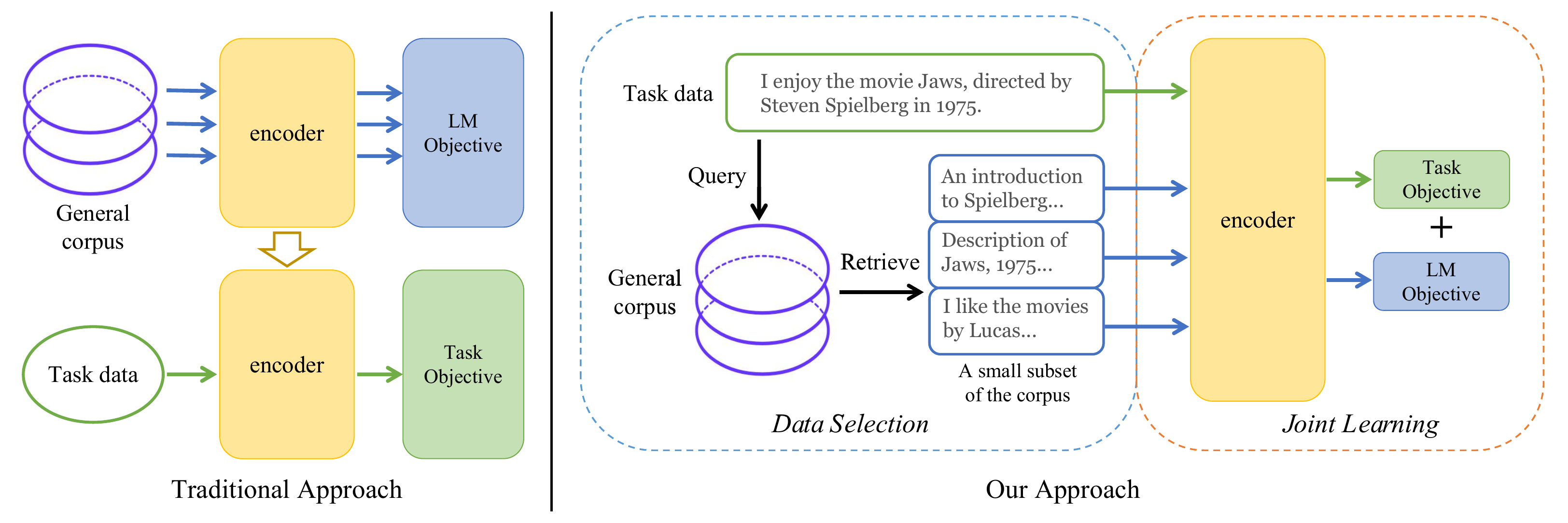}
    \vspace{-15pt}
    \caption{Comparison between the traditional pretraining-finetuning approach and our proposed framework \ours: instead of training a language model over the entire general corpus and then finetuning it on task data, we first use task data as queries to retrieve a tiny subset of the general corpus, and then perform joint learning on both the task objective and self-supervised language modeling objective.}
    \label{fig:overview}
\vspace{-10pt}
\end{figure*}

We evaluate \ours on eight different tasks covering the domains of news, review, computer science, and biomedical science, following the setting of~\citet{dontstoppretraining}. \ours achieves results better than or similar to BERT \cite{devlin2018bert} and RoBERTa \cite{liu2019roberta} while reducing the training FLOPs by \textbf{two orders of magnitude}\footnote{This effectively reduces the cost from training on 1,000 GPUs for one day to training on 8 GPUs for 42 hours.}.


\section{Related work}

\paragraph{Pretrained Language Models} 
Pretrained language models have become the de-facto solution to many of the NLP tasks~\cite{radford2018gpt,devlin2018bert,liu2019roberta,raffel2019exploring,brown2020language,yang2020xlnet}. 
Those models are usually pretrained on a large-scale corpus in a self-supervised manner to learn a contextualized representation of tokens in natural language, and then are fine-tuned with labeled data for specific tasks. 
BERT~\cite{devlin2018bert}, one of the most popular PLMs, is pretrained on a 16GB English corpus using a masked language modeling objective (i.e. predicting randomly masked tokens). 
RoBERTa~\cite{liu2019roberta} inherits the training objective of BERT, but is pretrained on a larger corpus consisting of 160GB English texts with larger batch size and dynamic token masking. 
In this work, we take both BERT and RoBERTa as our major baselines. 

\vspace{-10pt}
\paragraph{Efficient Pretraining for NLP}
There is a line of work dedicated to improving the efficiency of pretraining language models. 
\citet{you2019large} and \citet{shoeybi2019megatronlm} utilized the data and model parallelism across different computational devices to accelerate the pretraining process. However, accelerating through parallelism does not actually reduce computational costs in terms of FLOPs for training models at large scale. 
\citet{Chen_2021} and \citet{so2021primer} tried to identify efficient neural network architectures for language model pretraining, based on the lottery ticket hypothesis and neural architecture search. Such modifications on architecture can bring about $50\%\sim70\%$ reduction in computational costs.
\citet{clark2020electra} and \citet{he2020deberta} incorporated manually designed mechanisms into language model pretraining, such as adversarial training and disentangled representation of content and position, which brings about $50\%\sim75\%$ reduction in computational costs. \citet{gu2020train} proposed to use task-guided pre-training with selective masking, which reduces the computation cost by around 50\%.
In this work, orthogonal to the aforementioned works, we investigate improving efficiency by reducing training data redundancy. Our approach also results in more drastic improvements. 

\paragraph{Efficient Inference of Pretrained Models}
Another line of work aims at improving inference efficiency of PLMs.
Some works improve inference efficiency by distilling large PLMs into small-sized models and using the distilled models for inference, such as DistilBERT~\cite{sanh2019distilbert}, TinyBERT~\cite{jiao2019tinybert},
MobileBERT~\cite{MobileBERT},
FastBERT~\cite{FastBERT},
BORT~\cite{optimalsubarchitecture}, and
BERT-of-Theseus~\cite{BERT-of-Theseus}.
Other works speed up inference by quantizing PLMs with low-precision representations during inference, such as 
Q8-BERT~\cite{Q8BERT},
Q-BERT~\cite{QBERT}, and
I-BERT~\cite{IBERT}.
Another type of works, such as~\cite{sixtneeneheads,structuredpruning,weightpruning}, adopt pruning by removing parts of PLMs to make it smaller and faster.
However, these methods rely on large PLMs, and the performance after distillation, pruning, or quantization often decreases to a certain extent compared with some of the best PLMs (e.g., RoBERTa-Large).
In contrast, our approach doesn't rely on large-scale pre-training and achieves better or at least comparable performance.

\paragraph{Domain and Task Adaptation for Pretrained Models}

Domain-adaptive finetuning is a method that finetunes a pretrained model on in-domain data using a language modeling objective. It has been shown to be effective for domain and task adaptation~\cite{Zhang_2019,dontstoppretraining,li2020task,lee2020biobert}. There are a few crucial differences between domain-adaptive finetuning and \ours. First, \ours is a general method to improve training efficiency that does not use any additional domain data. It only utilizes the general corpus as in BERT and RoBERTa. In comparison, domain-adaptive finetuning uses domain data to improve domain adaptation. Second, while previous works on domain-adaptive finetuning are built upon a model pretrained on the general corpus, \ours learns from scratch without large-scale pretraining to substantially save computation costs.
\paragraph{Co-training for Semi-supervised Learning and Data-Density-Based Active Learning}
Additionally, we observe two techniques related to TLM. They are Co-Training (CT)~\cite{cotraining1,cotraining2} and Data-Density-Based Active Learning (DAL)~\cite{al1,al2} respectively.
Both CT and TLM utilize unlabeled data to aid the learning on a certain task. The difference between TLM and CT is 2-fold: First, CT requires training distinct models from multiple views of unlabeled data, yet TLM only trains a single model through pre-text tasks such as MLM. Second, TLM takes the selection process of unlabeled data into account, which is little discussed in CT. TLM and DAL share the same flavor of finding representative instances in a pool of unlabeled data. However, DAL makes the assumption that every unlabeled sample can be effectively labeled by the definition of the task, which is not required by TLM. Also, DAL tries to find critical instances iteratively from the whole pool of unlabeled data, yet TLM only tries to find relevant instances in a one-shot way with respect to labeled data, which makes TLM more efficient than classic DAL algorithms.





\section{Method}

\subsection{\ours: Task-Driven Language Modeling}
It is an interesting phenomenon that humans are able to quickly master a certain task with limited time and effort by focusing only on pieces of relevant knowledge. For example, when students cram for exams, they review a few chapters instead of going through all books in the world. 
Following this observation, we conjecture that one of the key aspects of learning a task is to quickly and precisely locate task-relevant information.
To this end, we develop \ours that first automatically retrieves relevant training data from a general corpus and then learns on the retrieved data and task data combined.

Formally, given a general corpus $\mathcal{D} = \{d_i\}_i$ where $d_i$ is a document, and labeled task data $\mathcal{T} = \{(x_i, y_i)\}_i$ where $x_i$ is text and $y_i \in \mathcal{Y}$ is a label\footnote{While it is straightforward to extend our framework to generation tasks, we focus on classification tasks in this work.}, our goal is to train a model $f$ to estimate the conditional probability for classification $f(x) = \hat{p}(y | x)$. 

\ours consists of two steps as shown in Figure~\ref{fig:overview}.
\begin{enumerate}[itemsep=2pt,topsep=0pt,parsep=0pt]
    \item Retrieve data from a general corpus using task data as queries.
    \item Train a model from scratch by jointly optimizing the task objective and the language modeling objective on the retrieved data and task data.
\end{enumerate}


\paragraph{Retrieval From General Corpus}
For each example in the task data $x_i \in \mathcal{T}$, we retrieve a set of documents $\mathcal{S}_i = \{\Tilde{d}_{i, 1}, \Tilde{d}_{i, 2}, \cdots\}$ from the given general corpus $\mathcal{D}$.
The set $\mathcal{S}_i$ represents the top-$K$ similar documents to $x_i$ in $\mathcal{D}$.
Retrieved data for all examples $x_i$ are combined $\mathcal{S} = \cup_i S_i$. Retrieved data $\mathcal{S}$ is a tiny subset of the general corpus $\mathcal{D}$.

We use BM25 \cite{bm25paper} for retrieval due to its efficiency. While using embedding-based dense retrievers \cite{karpukhin2020dense} might lead to better retrieval results, we do not consider these methods to keep our approach as simple as possible. Moreover, dense retrievers rely on pretraining, which might bring additional computational costs. The exploration of achieving a better tradeoff between efficiency and retrieval performance is left to future work.
Moreover, for tasks with extremely long texts (e.g., Helpfulness~\cite{helpfulness}), we find it more efficient to extract keywords (e.g., using the RAKE algorithm \cite{rose2010automatic}) to form the queries for retrieval instead of using the entire input sequence.
We call the retrieved data $\mathcal{S}$ external data and the task data $\mathcal{T}$ internal data.

Note that our data retrieval method is task-agnostic---it only depends on text $x$ without dependency on $y$. Moreover, the retrieval procedure does not assume the availability of domain-specific data. It operates on a general corpus and has the same input as the pretraining-finetuning paradigm.

\paragraph{Joint Training}
Given both the internal and external data, we train a language model $f$ from scratch.
Let $\mathcal{L}_\text{mlm}(x)$ be the masked language modeling loss as in BERT \cite{devlin2018bert}, and let $\mathcal{L}_\text{task}(f(x), y)$ be the task loss function (e.g., cross entropy for classification). \ours optimizes the following loss function:
\[
\begin{split}
&\rho_1\mathbb{E}_{x \sim \mathcal{S}} [\mathcal{L}_\text{mlm}(x)] \\
+ &\mathbb{E}_{x, y \sim \mathcal{T}} \left[ \rho_2 \mathcal{L}_\text{mlm}(x) + \mathcal{L}_\text{task}(f(x), y) \right]
\end{split}
\]
where $\rho_1$ and $\rho_2$ are hyperparameters. The network architecture we employ is identical to BERT, where we use a CLS head for classification and an LM head for masked language modeling. \ours can also be extended to other architectures for non-classification tasks. Our implementation involves a two-stage training procedure. In the first stage, we interleave one batch of internal data with $\rho_1$ batches of external data for mini-batch stochastic gradient descent, where $\rho_1$ is set as an integer. In the second stage, we set both $\rho_1$ and $\rho_2$ as zero to only finetune the model on internal data with the task objective.




\subsection{Comparison Between \ours and PLMs} \label{sec:tlmplm}

\begin{table}
    \centering
    \caption{Comparison between \ours and PLMs. Here we provide qualitative comparison, while quantitative comparison in terms of training data size, FLOPs, and the number of parameters is available in Table~\ref{tab:mainresults}.
    }
    \vspace{0.33cm}
    \resizebox{0.5\textwidth}{!}{%
    \begin{tabular}{l|c|c}
    \toprule[1pt]
    & \tf{\ours} & \tf{PLMs} \\
    \midrule
    Loss Function & $\mathcal{L}_{task}$ and $\mathcal{L}_{mlm}$ & $\mathcal{L}_{mlm}$  \\ \hline
    Training Data & \makecell{A tiny subset of $D$ and task data $\mathcal{T}$} & The entire $D$  \\\hline
    Compute Cost & \makecell{8 GPUs \\ 42 hours} & \makecell{1,000 GPUs \\ one day} \\ \hline
    Generality & Task-Driven & Task-Agnostic \\
    \bottomrule[1pt]
    \end{tabular}
    }
    \label{tab:comparisonofmethod}
    \vspace{-5pt}
\end{table}
Both \ours and pretraining-finetuning have two stages. In fact, the second stage of \ours equals the traditional finetuning stage.
The main difference between the first stage of \ours and pretraining (PLMs) is shown in Table~\ref{tab:comparisonofmethod}. Unlike PLMs which learn as much task-agnostic knowledge as possible at an extremely high cost, \ours learns task-related knowledge for each task with very low costs.

Given the above difference between \ours and PLMs, we will discuss the pros and cons of \ours in detail.





\paragraph{Democratizing NLP}
In pretraining-finetuning paradigm, the finetuning performance is largely upper bounded by the pretrained model. However, due to the constraints of computational resources, the majority of NLP researchers cannot afford training large-scale language models and resort to studying the finetuning algorithms. Since only a small portion of researchers are working on the architectures, loss functions, and other design choices of PLMs, there is a risk that the development of the field might be slowing down. On the other hand, \ours is efficient and highly performant. As a result, \ours has the potential of democratizing NLP and expediting its development by allowing most researchers to freely explore the architectures, loss functions, algorithms, and other design choices in the neighborhood of a state-of-the-art solution.

\paragraph{Efficiency}
\ours improves over PLMs in terms of per-task FLOPs. In many cases when there are only a few target tasks, \ours is favorable. For example, a researcher might be interested in solving four textual entailment datasets, or an industrial team might want to improve a recommender system which can be viewed as one task. However, if the goal is to solve 1,000 tasks at once (e.g., building an NLP platform to serve multiple business units within a corporate), PLMs might still be preferred.

\paragraph{Flexibility}
Since \ours is task-driven, there is a larger degree of flexibility. Researchers can use custom strategies for tokenization, sequence length, data representations, hyperparameter tuning, etc, which might improve performance and/or efficiency.

\paragraph{Generality}
PLMs learn task-agnostic general representations and can be used for few-shot and zero-shot learning \cite{brown2020language}. In comparison, \ours trades generality for efficiency by learning only task-specific representations. How to further improve \ours in terms of learning more general representations poses a challenge for future work. We believe multi-task learning might alleviate this issue given recent observations \cite{wei2021finetuned,leeadapting}, especially for in-domain zero-shot generalization. It might also be possible to combine pretraining with \ours, e.g., using a small PLM with \ours to match a larger PLM, to achieve a better tradeoff between generality and efficiency.

\section{Experiments}

\begin{table*}[ht]
\caption{Evaluation results for \ours at three different training scales. For each task, we report the average F1 score across three random seeds with standard deviations as subscripts. We also list the number of parameters, the total training compute (FLOPs), and the size of training corpus for comparison.}
\vspace{-10pt}
\begin{center}
\centering
\resizebox{\textwidth}{!}{
\begin{threeparttable}
  \begin{tabular}{p{2.6cm}|p{1cm}<{\centering}p{1.1cm}<{\centering}p{1.2cm}<{\centering}|ccccccccc}
    \toprule[1pt]
    \tf{Model} & \tf{\#Param} &  \tf{FLOPs}$^1$ & \tf{Data}$^2$ & \tf{AGNews} & \tf{Hyp.} & \tf{Help.} & \tf{IMDB} & \tf{ACL.} & \tf{SciERC} & \tf{Chem.} & \tf{RCT} & \tf{Avg.} \\
    \midrule[1pt]
    \multirow{2}*{BERT-Base$^3$} &
    \multirow{2}*{109M} & \multirow{2}*{2.79E19} & \multirow{2}*{16GB}
    & 93.50 &	91.93&	69.11&	93.77&	69.45&	80.98 &	81.94&	87.00&	\multirow{2}*{83.46} \\
    &&&& \std0.15 & \std1.74 & \std0.17 & \std0.22 & \std2.90 & \std1.07 & \std0.38 & \std0.06 & \\
    \\[-2.0ex]
    \multirow{2}*{BERT-Large$^3$} &
    \multirow{2}*{355M} & \multirow{2}*{9.07E19} & \multirow{2}*{16GB}
    & 93.51&	91.62 &	69.39&	\tf{94.76}&	69.13&	\tf{81.37}&	\tf{83.64}&	\tf{87.13}&	\multirow{2}*{\tf{83.82}}\\
    &&&& \std0.40 & \std0.69 & \std1.14 & \std0.09 & \std2.93 & \std1.35 & \std0.41 & \std0.09 & \\
    \\[-2.0ex]
    \name & 
    \multirow{2}*{109M} & \multirow{2}*{2.74E18} & \multirow{2}*{0.91GB}
    & \tf{93.74}&	\tf{93.53}&	\tf{70.54}&	93.08&	\tf{69.84}&	80.51&	81.99&	86.99&	\multirow{2}*{83.78} \\
    \textit{(small-scale)}&&&& \std0.20 & \std1.61 & \std0.39 & \std0.17 & \std3.69 & \std1.53 & \std0.42 & \std0.03 & \\
    \midrule[1pt]
    \multirow{2}*{RoBERTa-Base$^3$} & \multirow{2}*{125M} & \multirow{2}*{1.54E21} & \multirow{2}*{160GB}
    & \tf{94.02}&	93.53&	70.45&	\tf{95.43}&	68.34&	81.35&	82.60 &	87.23&	\multirow{2}*{84.12} \\
    &&&& \std0.15 & \std1.61 & \std0.24 & \std0.16 & \std7.27 & \std0.63 & \std0.53 & \std0.09 & \\
    \\[-2.0ex]
    \name & 
    \multirow{2}*{109M} & \multirow{2}*{8.30E18} & \multirow{2}*{1.21GB}
    & 93.96&	\tf{94.05}&	\tf{70.90}&	93.97&	\tf{72.37}&	\tf{81.88}&	\tf{83.24}&	\tf{87.28}&	\multirow{2}*{\tf{84.71}} \\
    \textit{(medium-scale)} &&&& \std0.18 & \std0.96 & \std0.73 & \std0.10 & \std2.11 & \std1.92 & \std0.36 & \std0.10 & \\
    \midrule[1pt]
    \multirow{2}*{RoBERTa-Large$^3$} & \multirow{2}*{355M} & \multirow{2}*{4.36E21} & \multirow{2}*{160GB}
    & 94.30 & \tf{95.16} & 70.73 & \tf{96.20} & \tf{72.80} & 82.62 & 84.62 & \tf{87.53} & \multirow{2}*{85.50} \\
    &&&&\std0.23 & \std0.00 & \std0.62 & \std0.19 & \std0.62 & \std0.68 & \std0.50 & \std0.13 & \\
    \\[-2.0ex]
    \name & \multirow{2}*{355M} & \multirow{2}*{7.59E19} & \multirow{2}*{3.64GB}
    & \tf{94.34} & \tf{95.16} & \tf{72.49} & 95.77 & 72.19 & \tf{83.29} & \tf{85.12} & 87.50 & \multirow{2}*{\tf{85.74}} \\
    \textit{(large-scale)}&&&& \std0.12 & \std0.00 & \std0.33 & \std0.24 & \std1.72 & \std0.95 & \std0.85 & \std0.12 & \\
    \bottomrule[1pt]
    \end{tabular}
\begin{tablenotes}
     \item[1] The total training compute (FLOPs) is calculated by $(6 \times \text{Total\_Training\_Tokens} \times \text{Parameter\_Size})$ as in~\cite{brown2020language}. For \ours, FLOPs are reported as the averaged result over eight tasks.
     \item[2] The size of data selected from general corpus that are actually used in training. For \ours, it is reported by averaging over eight tasks.
     \item[3] The BERT-Base and BERT-Large are pretrained by~\cite{devlin2018bert} and RoBERTa-Base and RoBERTa-Large are pretrained by~\cite{liu2019roberta}. We finetuned them to obtain the results over the eight tasks.
\end{tablenotes}
\end{threeparttable}
}
\end{center}
\vspace{-15pt}
\label{tab:mainresults}
\end{table*}




\subsection{Setup}

\paragraph{Datasets}
Following~\cite{dontstoppretraining},
we conduct experiments on eight tasks over four domains, including biomedical science, computer science, news, and reviews (two tasks in each domain).
The tasks can be categorized into high-resource and low-resource tasks.
High-resource tasks has more than 5K task data, including AGNews~\cite{agnews}, IMDB~\cite{imdb}, RCT~\cite{RCT}, and Helpfulness~\cite{helpfulness}, while low-resource tasks include ChemProt~\cite{chemprot}, ACL-ARC~\cite{ACLARC}, SciERC~\cite{SCIERC}, and HyperPartisan~\cite{hyperpartisan}.
For the general training corpus, we collected two corpora that respectively match the original training corpora of BERT and RoBERTa.
We name them respectively Corpus-BERT (\bertcorpus) and Corpus-RoBERTa (\robertacorpus).
The size of \robertacorpus is 10 times larger than \bertcorpus.

\vspace{-5pt}
\paragraph{Baselines}
Our experiments focus on comparison with general PLMs.
We finetuned both BERT~\cite{devlin2018bert} and RoBERTa~\cite{liu2019roberta} of base and large scales as the baselines.
Although \ours is a general method without using addition in-domain data, it even performs close to domain-adaptive finetuning methods \cite{dontstoppretraining} (see Appendix \ref{sec:dapt_compare} for detailed comparison).

\paragraph{Evaluation Strategy}
We report the average performance across three random seeds, together with the standard deviation.
We follow ~\citet{scibert} and ~\citet{dontstoppretraining} to report the test micro-F1 for ChemProt and RCT, and macro-F1 for the rest of the datasets.

For fair comparison, we evaluate \ours of different \textit{training scales}.
The training scale is defined by three factors, including the number of parameters, the size of the general corpus, and the number of total training tokens.
The number of total training tokens is calculated as the product of training steps, batch size, and sequence length.
We report \ours at three training scales as shown in Table~\ref{tab:computation}, namely \textit{small}, \textit{medium}, and \textit{large} scales.
Each scale of \ours is accordingly compared to the PLM baselines with an increasing computational cost.

\paragraph{Training Details}
For each experiment of \ours, while fixing the training scale hyper-parameters (i.e., training steps, batch size and sequence length), we perform a grid search over \eir and \mw.
We listed the hyper-parameters used in Table~\ref{tab:computation} in Appendix.

\subsection{Main Results} \label{sec:main_results}

Table~\ref{tab:mainresults} shows the main results that compare \ours of three different scales and the according PLM baselines.
In conclusion, \ours can achieve results that are better than or comparable to the baselines with substantial reduction in FLOPs and the size of training data.
Specifically, at a small scale, \ours achieves comparable results to BERT-Large with an average of 1/33 of FLOPs and 1/16 of the training corpus.
At the medium and large scales, \ours improves the performance by $0.59$ and $0.24$ points on average respectively, while significantly reducing both FLOPs and the training data size by two orders of magnitude or more.
These results confirm that \ours is highly accurate and much more efficient than PLMs.
Moreover, \ours gains more advantages in efficiency at a larger scale. This indicates that larger-scale PLMs might have been trained to store more general knowledge that is not useful for a specific task.

\subsection{Ablation Study}


\subsubsection{Data Retrieval}\label{sec:dataselectionablation}

Table~\ref{tab:data_quality} shows the comparison between different retrieval methods (i.e., BM25 and random retrieval) and different sizes of the general corpus.
We find that given the same general corpus, the results of BM25 significantly outperform those of random retrieval by a large margin on all tasks, showing that using task-relevant data for joint training is crucial for the best performance.
Specifically, BM25 shows an advantage of almost 1 point against random retrieval on high-resource tasks such as IMDB, and more significant advantages on low-resource tasks such as SciERC and ChemProt by around 3-4 points. This is aligned with our intuition that low-resource tasks rely more on external data.

\begin{table}[ht]
\vspace{-10pt}
\caption{Results on the development set using different retrieval methods and different general corpora on each task. We compared two data retrieval methods: random retrieval and the BM25 algorithm. We compare two source general corpora: the corpus used in BERT (\bertcorpus) and the corpus used in RoBERTa (\robertacorpus). The size of \robertacorpus is 10 times larger than \bertcorpus.}
\vspace{-10pt}
    \begin{center}
    \centering
    \resizebox{\columnwidth}{!}{%
      \begin{tabular}{lccc}
        \toprule[1pt]
          & \tf{IMDB} & \tf{SciERC} & \tf{ChemProt} \\
        \midrule
        Random &\\
        ~~w/ \bertcorpus &
        93.65\std{0.09}&83.80\std{0.62}&80.65\std{0.48} \\
        ~~w/ \robertacorpus & 
        94.04\std{0.22}&83.10\std{1.54} &80.73\std{0.46}\\
        BM25 &\\
        ~~w/ \bertcorpus &
        94.40\std{0.09} &86.07\std{0.48} & 83.64\std{0.26}\\
        ~~w/ \robertacorpus &
        \tf{94.90}\std{0.06} & \tf{87.41}\std{0.36}& \tf{84.99}\std{0.72}\\
        \bottomrule[1pt]
        \end{tabular}
        }
\end{center}
\vspace{-10pt}
\label{tab:data_quality}
\end{table}

By comparing the results of \bertcorpus and \robertacorpus with BM25, we observe that increasing the size of the general corpus improves performance (by 0.5, 1.34, and 1.35 points on IMDB, SciREC, and ChemProt respectively). The gains of using 10 times more data are similar to the ones observed in PLMs \cite{yang2020xlnet,liu2019roberta}. This indicates that although \ours only uses a small amount of data, it is able to scale when a larger general corpus is available while maintaining efficiency. On the other hand, the gains of using a larger corpus diminish with random retrieval, showing that random retrieval, as a task-agnostic method, is not very sensitive to the general corpus size.

\begin{table}[ht]
\caption{Results on the development set with different values of $K$. The value $K$ is the number of retrieved documents per task example. AGNews is a high-resource task, while SciREC and ChemProt are low-resource ones. Here we use \mw = 20 for all tasks. When there are external data available, we use \eir = 4 for AGNews and \eir = 1000 for SciERC and ChemProt.}
\vspace{-10pt}
    \begin{center}
    \centering
    \resizebox{\columnwidth}{!}{%
      \begin{tabular}{lccc}
        \toprule[1pt]
          & \tf{AGNews} & \tf{SciERC} & \tf{ChemProt} \\
        \midrule
        Only Task Data& 93.41\std{0.10} & 51.23\std{1.13} & 55.05\std{0.18} \\
        Top-50& \tf{94.51}\std{0.15}&77.61\std{1.75}& 77.21\std{0.47} \\
        Top-500& 94.32\std{0.05}& 82.39\std{0.55}  & 81.44\std{0.50}\\
        Top-5000& 94.42\std{0.10} & \tf{86.07}\std{0.48} & \tf{83.64}\std{0.26}\\
        \bottomrule[1pt]
        \end{tabular}
        }
      \end{center}
    \vspace{-10pt}
    \label{tab:data_amount}
\end{table}


Data retrieval selects the top-$K$ similar documents from the general corpus.
Table~\ref{tab:data_amount} shows the results of different $K$ values.
We observe that high-resource tasks such as AGNews only need a small $K$ value, while low-resource tasks such as SciREC and ChemProt require a large $K$ to obtain the best performance.
The observation is consistent with the above analysis that low-resource tasks rely more on external data to improve from joint training.


\begin{figure*}
\centering
\subfigure[\ours (\textit{Medium scale})]{
\begin{minipage}[t]{0.33\linewidth}
\centering
\includegraphics[width=2in]{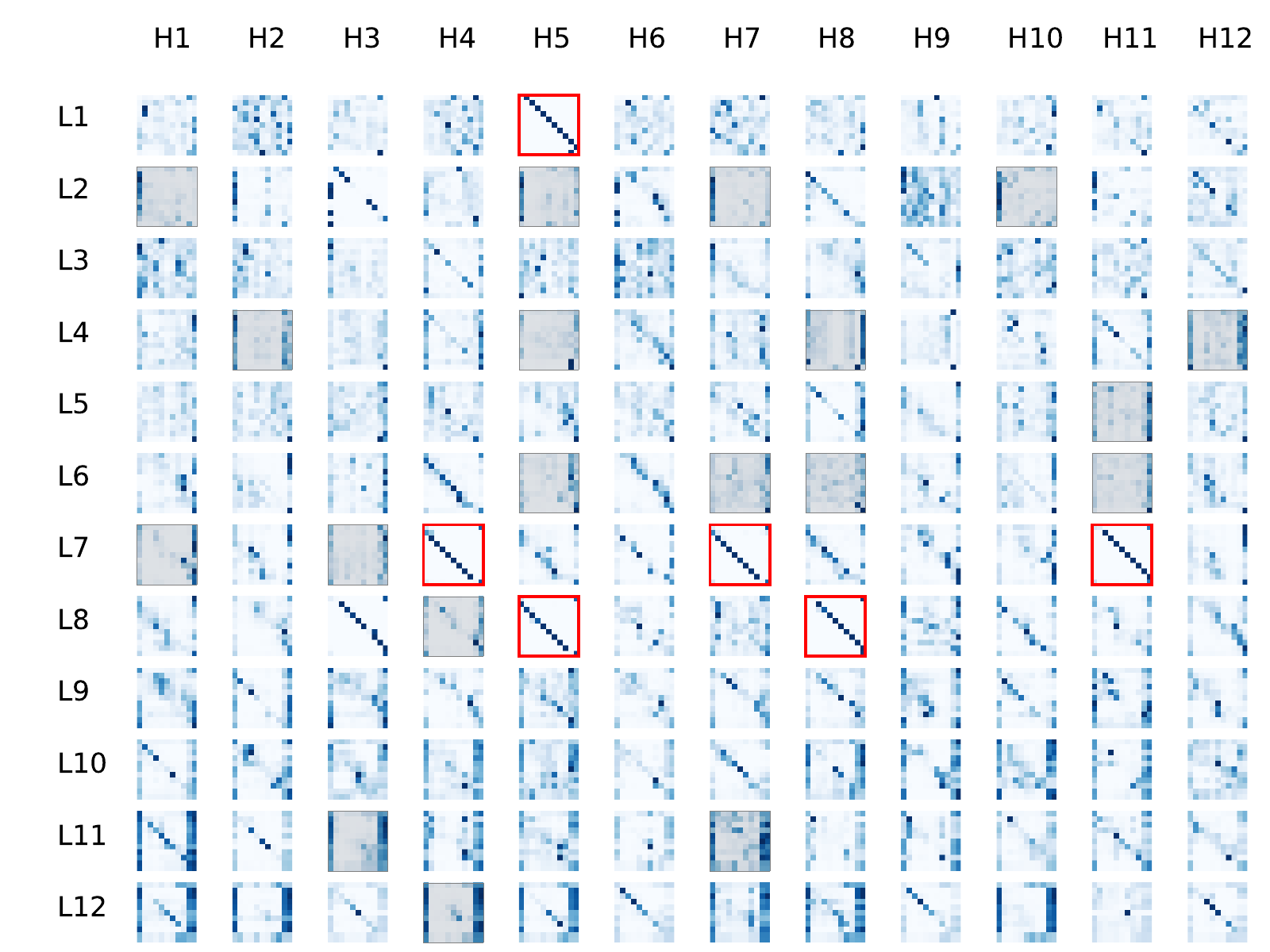}
\end{minipage}%
}%
\subfigure[BERT-Base]{
\begin{minipage}[t]{0.33\linewidth}
\centering
\includegraphics[width=2in]{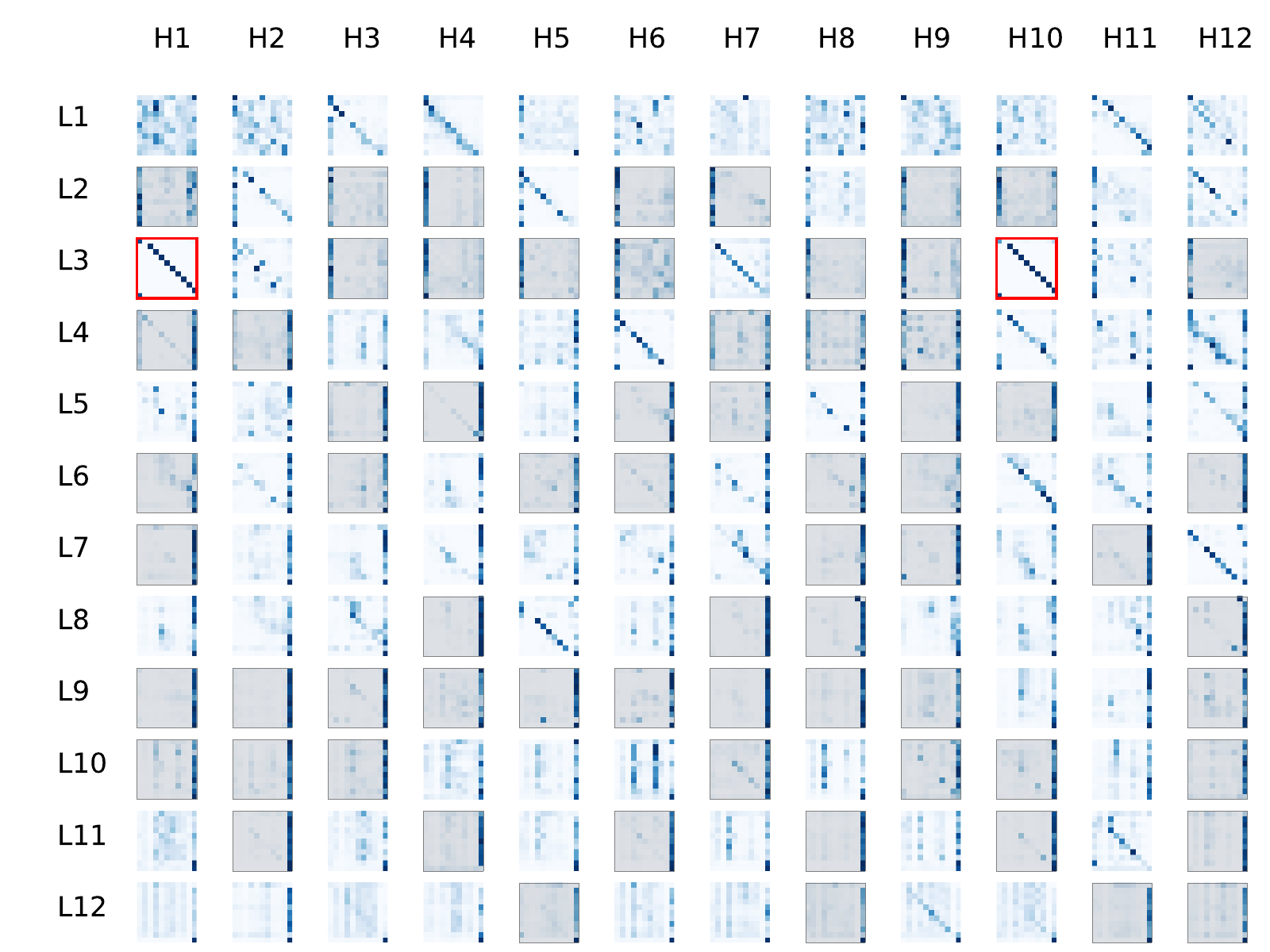}
\end{minipage}%
}%
\subfigure[RoBERTa-Base]{
\begin{minipage}[t]{0.33\linewidth}
\centering
\includegraphics[width=2in]{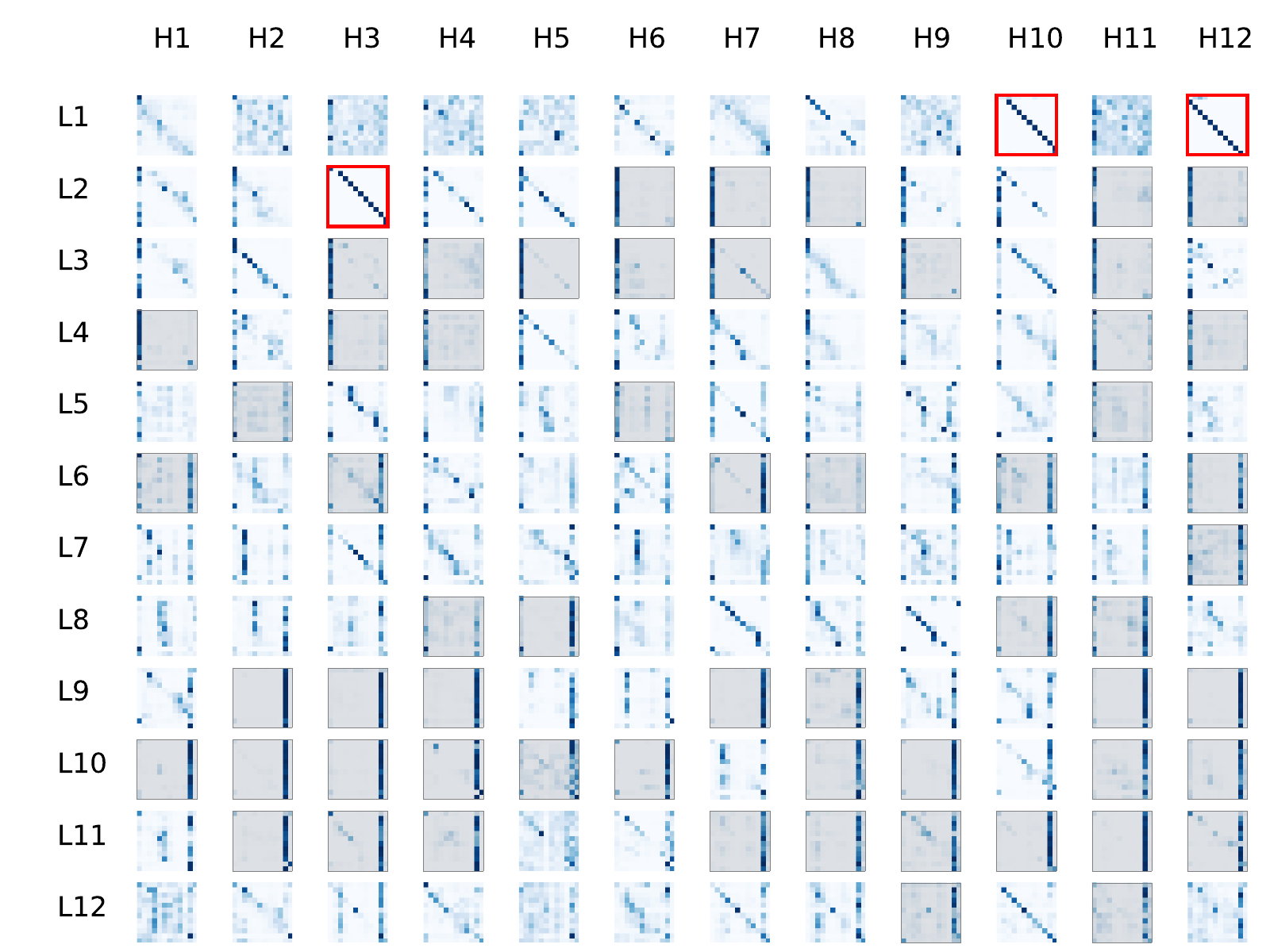}
\end{minipage}
}%
\vspace{-10pt}
\caption{Attention visualization of \ours and pretraining-finetuning baselines, with "[CLS] crystallographic comparison with the structurally related. [SEP]" from ChemProt as the input. 
The positional heads~\cite{voita-etal-2019-analyzing} are highlighted in red boxes and vertical heads~\cite{kovaleva-etal-2019-revealing} are masked in gray.}
\vspace{-10pt}
\label{visual}
\end{figure*}

\subsubsection{Language Modeling Weights \eir and \mw}\label{sec:ablationsentimentanalysis}

The hyperparameters \eir and \mw are the weights for the LM loss on external and internal data respectively.
We conduct sensitivity analysis over \eir and \mw. 
Results are shown in Table~\ref{tab:external_ratio} and Table~\ref{tab:lm_weight}.

\begin{table}[ht]
\small
\vspace{-10pt}
\caption{Results on the development set with different weights on external data (i.e., \eir). 
        We assign different values for \eir for the first stage, and report the final performance after two-stage joint learning.
        ``Ext only'' means using only external data for training (i.e., $\rho_1 = \infty$).
        Helpfulness is a high-resource task, and the others are low-resource ones. For all tasks, we fix \mw = 20.}
    \begin{center}
    \centering
      \begin{tabular}{lccc}
        \toprule[1pt]
         & \tf{Helpfulness} & \tf{SciERC} & \tf{ChemProt} \\
        \midrule
        \eir=1 & \tf{71.02}\std{0.51} & 80.72\std{3.32} & 73.27\std{0.30}\\
        \eir=3 & 70.41\std{0.52} & 80.01\std{0.72} & 79.43\std{1.03} \\
        \eir=99 &69.56\std{0.23} & 84.95\std{0.57} & 83.30\std{0.30}\\ 
        \eir=999 &69.35\std{0.72} & \tf{86.07}\std{0.48} & \tf{83.64}\std{0.26}\\
        Ext only & 69.76\std{0.50} & 85.66\std{1.58} & 82.50\std{0.27}\\
        \bottomrule[1pt]
        \end{tabular}
      \end{center}
      \vspace{-10pt}
        \label{tab:external_ratio}
\end{table}
\begin{table}[ht]
\vspace{-10pt}
\caption{Results on the development set with different language modeling weights on internal data (i.e., \mw). Here we set \eir = 1000 for SciERC and ChemProt, and \eir = 4 for RCT}
\small
    \begin{center}
    \centering
      \begin{tabular}{lcccc}
        \toprule[1pt]
         & \tf{RCT} & \tf{SciERC} & \tf{ChemProt} \\
        \midrule
        \mw=0 & 85.75\std{0.11} & 83.31\std{0.88} & 83.41\std{0.33}\\
        \mw=20 & 88.08\std{0.02} & \tf{86.07}\std{0.48} & 83.64\std{0.26}\\
        \mw=100 & \tf{88.16}\std{0.15} & 85.48\std{1.01} & \tf{83.77}\std{0.77}\\
        \mw=1000 &88.02\std{0.04} & 85.29\std{1.86} & 83.63\std{0.90}\\ 
        \bottomrule[1pt]
        \end{tabular}
      \end{center}
      \vspace{-10pt}
        \label{tab:lm_weight}
\end{table}

For \eir, we find that high-resource tasks such as Helpfulness perform better with a smaller \eir (i.e., Helpfulness achieves best when \eir$=1$) while low-resource tasks such as SciERC and ChemProt achieve their best when \eir is large (i.e., both tasks use \eir$=999$).
This is in line with conclusions in Section~\ref{sec:dataselectionablation} that low-resource tasks rely more on external data.
In addition, removing task data and only using external data for training (i.e., \eir=\#\bertcorpus), it performs worse than when incorporating the task data, proving the indispensability of small task data.

Results in Table~\ref{tab:lm_weight} show that language modeling on internal data is necessary: consistently better results are achieved when \mw is non-zero. Based on our observations, competitive performance can be achieved when \mw is set to a proper value between 20 and 1000.


\subsubsection{Second Stage of Training}

\ours contains two training stages---first training on all three terms combined and then finetuning using only the task objective.
To validate the effectiveness of the second stage of \ours, we compare the performance of two-stage training against using only stage one.
Results are shown in Table~\ref{tab:stage_compare}.
We find that removing the second stage hurts the ultimate performance consistently, proving its indispensability.
Particularly, the second stage has much more influence on low-resource tasks (with a huge decrease of 19.37 points on ACL-ARC and 14.34 points on ChemProt) than on high-resource tasks (with a performance decrease of 0.53 points on AGNews and 2.17 points on IMDB).

\begin{table}[ht]
\vspace{-15pt}
\caption{
        Results on the development set of two-stage training and one-stage training (removing stage 2).
        }
    \vspace{-10pt}
    \begin{center}
    \centering
    \resizebox{\linewidth}{!}{%
      \begin{tabular}{lcccc}
        \toprule[1pt]
         & \tf{AGNews} & \tf{IMDB} & \tf{ChemProt} & \tf{ACL-ARC}\\
        \midrule
        two-stage & \tf{94.51} &  \tf{94.40} &\tf{83.64}& \tf{76.37}\\
        wo/ stage-2 & 93.98 {\color{red}$\downarrow$} & 92.23{\color{red}$\downarrow$} & 69.30{\color{red}$\downarrow$} & 57.00{\color{red}$\downarrow$}\\
        \bottomrule[1pt]
        \end{tabular}
        }
      \end{center}
        \label{tab:stage_compare}
        \vspace{-10pt}
\end{table}

\subsubsection{MLM Loss on Task Data}
During the first training stage, TLM uses masked language loss on task data. To examine whether the trick attains the main improvements, we compare results on PLM, PLM with additional MLM loss on task data (PLM+MLM) and TLM. Results in Table~\ref{tab:mlmplm} show that adding MLM loss on task data into PLM has
only marginal gains and does not affect the main
conclusion of the paper. 
In addition,
results in Table~\ref{tab:data_quality} and Table~\ref{tab:data_amount} show that retrieving appropriate relevant data is also essential for the
performance of TLM. 

\begin{table}
\vspace{-15pt}
\caption{\small
Results of adding MLM loss on task data into PLM. Results are based on RoBERTa-base.
}
\vspace{5pt}
\centering
\small
\resizebox{0.5\textwidth}{!}{
  \begin{tabular}{l|ccccccccc}
    \toprule[1pt]
    \textbf{Model} &
    \textbf{AGNews} & \textbf{Hyp.} & \textbf{Help.} & \textbf{IMDB} & \textbf{ACL.} & \textbf{SciERC} & \textbf{Chem.} & \textbf{RCT} & \textbf{Avg.} \\
    \midrule[1pt]
    PLM &
    94.02 &
    93.53 &
    70.45 &
    95.43 & 
    68.34 &
    81.35 &
    82.60 &
    87.23 & 84.12 \\
    PLM+MLM &
    93.83 &
    93.50 & 
    71.12 &
    95.54 &
    70.94 &
    80.90 &
    82.53 &
    87.09 & 
    84.43 \\
    TLM  &
    93.96 &	
    94.05 &
    70.90 &	
    93.97 &
    72.37 &
    81.88 &
    83.24 &
    87.28 &
    84.71 \\
    \bottomrule[1pt]
    \end{tabular}
    }
\label{tab:mlmplm}
\vspace{-15pt}
\end{table}

\begin{table*}[!h]
\caption{Examples of retrieved data. The overlap between queries and retrieved data are highlighted in blue in italics.}
\vspace{0.3cm}
\begin{center}
\small
\centering
\begin{tabular}{p{1.2cm}|l|l}
\toprule[1pt]
\tf{Task} & \tf{Task Data as Query} & \tf{Retrieved General Data} \\
\midrule

    \begin{tabular}[c]{@{}c@{}} Hyp. \end{tabular}
    & 
    \begin{tabular}[c]{p{0.3\textwidth}}
    "A Republican student association at \hl{San Diego State University (SDSU)} is facing backlash for sending a letter demanding Muslim students condemn last week’s terror attacks in Barcelona. ... " \\
    \end{tabular}
    & 
    \begin{tabular}[c]{p{0.5\textwidth}}
    
        {\begin{tabular}[c]{p{0.5\textwidth}}
        \textbf{Example 1:} "...The \hl{SDSU} Aztecs intercollegiate water polo, swimming and diving teams are based at the Aztec Aquaplex..." \\
        \textbf{Example 2:} The Daily Aztec is a not-for-profit, independent student newspaper serving \hl{San Diego State University (SDSU)} and the surrounding College Area in San Diego, California. ...\\ 
        \end{tabular}}
         \\
    \end{tabular}\\
    \midrule
    \begin{tabular}[c]{@{}c@{}} Help. \end{tabular}
    & 
    \begin{tabular}[c]{p{0.3\textwidth}}
    \hl{Poor Quality}. The case broke after dropping it on the tile floor. ... \\
    \end{tabular}
    & 
    \begin{tabular}[c]{p{0.5\textwidth}}
    
        {\begin{tabular}[c]{p{0.5\textwidth}}
        \textbf{Example 1:} ...a collaborative algorithm will be able to recommend it, the \hl{quality} of those recommendations will be \hl{poor}. ... \\
        \textbf{Example 2:} ... Books that're of \hl{poor quality} will quickly cease to sell. ... \\
        \end{tabular}}
         \\
    \end{tabular}\\
    \midrule
    \begin{tabular}[c]{@{}c@{}} ChemProt \end{tabular}
    & 
    \begin{tabular}[c]{p{0.3\textwidth}}
    FCEO significantly inhibited nitric oxide (NO) and prostaglandin E2 (PGE2) by suppressing the protein expression of \hl{inducible nitric oxide synthase (iNOS)} and \hl{cyclooxygenase (COX)-2}, respectively. 
    \end{tabular}
    & 
    \begin{tabular}[c]{p{0.5\textwidth}}
    
        {\begin{tabular}[c]{p{0.5\textwidth}}
        \textbf{Example 1:} ... They regulate the development of sperm by controlling their cell division and survival. Other immune factors found in the testis include the enzyme \hl{inducible nitric oxide synthase (iNOS)} ...  \\
        \textbf{Example 2:} These compounds have been shown "in vivo" to reduce two proteins that mediate inflammation, \hl{cyclooxygenase-2 (COX-2)} and \hl{inducible nitric oxide synthase (iNOS)}. ... \\
        \end{tabular}}
         \\
    \end{tabular}\\
    \midrule
    \begin{tabular}[c]{@{}c@{}} SciERC \end{tabular}
    & 
    \begin{tabular}[c]{p{0.3\textwidth}}
    \hl{Image} sequence \hl{processing} techniques are used to study exchange , growth , and transport processes and to tackle key questions in environmental physics and biology. \\
    \end{tabular}
    & 
    \begin{tabular}[c]{p{0.5\textwidth}}
    
        {\begin{tabular}[c]{p{0.5\textwidth}}
        \textbf{Example 1:} ...  Driving forces in signal \hl{processing} for data parallelism are video encoding, \hl{image} and graphics \hl{processing}, wireless communications to name a few. \\
        \textbf{Example 2:} They have applications in many disciplines such as biology, chemistry, ecology, neuroscience, physics, \hl{image} \hl{processing}, ... \\ 
        \end{tabular}}
         \\
    \end{tabular}\\    
    
\bottomrule[1pt]
\end{tabular}
\vspace{-10pt}
\end{center}
\label{tab:retrieve}
\end{table*}

\begin{table*}
\caption{Evaluation results on the GLUE benchmark. Model size, data, and FLOPs are similar to Table \ref{tab:mainresults}.}
\small
\vspace{0.3cm}
    \begin{center}
    \centering
    \begin{threeparttable}
      \begin{tabular}{p{3cm}|ccccccccc}
        \toprule[1pt]
        Method &  
        \tf{CoLA} & 
        \tf{RTE} & 
        \tf{STS-B} & 
        \tf{MRPC} & 
        \tf{QQP} & 
        \tf{SST-2} & 
        \tf{QNLI} &
        \tf{MNLI} & 
        \tf{Avg.} \\
        \midrule[1pt]
BERT-Base &
59.3 &
68.2 &
89.8/89.4 & 
86.0/90.5 &
91.1/88.1 &
92.5 &
91.8 &
84.5/84.5 &
82.97 \\
TLM \textit{(small-scale)} &
59.8 & 
67.1 &
89.0/88.7 &
86.8/90.4 &
91.1/88.1 &
92.2 &
91.0 &
83.3/83.9 &
82.60 \\
        \bottomrule[1pt]
        \end{tabular}
\end{threeparttable}
\end{center}
\label{tab:glue}
\vspace{-10pt}
\end{table*}

\subsection{Analysis}

\subsubsection{Attention Weight Visualization}

We also study the difference between the model behaviors of \ours and pretraining-finetuning by visualizing their attention weights. 
\citet{voita-etal-2019-analyzing} found that a specific kind of heads, referred to as "positional head" in which at least 90\% of the maximum attention weights are assigned to adjacent tokens, have vital contributions to final predictions of the model.
Another sort of heads we are interested in are those in which most maximum attention weights are assigned to [CLS],[SEP] or the period token("."), which potentially encode less semantic or syntactic information \cite{kovaleva-etal-2019-revealing}. 
In our experiments, if more than 90\% maximum weights are assigned to [CLS], [SEP] or the period token, we categorize this head as a ``vertical head''.
Results in Figure \ref{visual} show that on the task ChemProt, more positional heads and less vertical heads are observed in \ours than in PLMs.
We also observe similar patterns across various tasks (see Appendix \ref{sec:more_vis}).
These phenomena suggest that TLM learns different (probably more informative) attention patterns compared to PLMs.

\subsubsection{Case Study of Retrieved Data}
We have shown several casess of retrieved data in Table~\ref{tab:retrieve}.
\ours retrieves relevant data from a general corpus using BM25 \cite{bm25paper}. 
Since BM25 is based on sparse features, it focuses more on lexical similarity instead of semantic similarity. 
This might be specifically beneficial for professional domains, e.g., SciERC for computer science and ChemProt for biomedical science), since there are a large number of proper nouns in these domains.
For other domains, it seems BM25 also performs reasonably well for retrieving related documents.

\subsection{Results on More Datasets}

So far we have followed the setting of \citet{dontstoppretraining} and adopted the datasets therein. 
In this section, we additionally experiment with the GLUE benchmark \cite{wang2018glue} following the setting of BERT \cite{devlin2018bert} to examine the performance of \ours on a more diverse set of tasks including natural language understanding. We follow the \textit{small-scale} setting in Section \ref{sec:main_results} in terms of model size, data, and FLOPs. Results in Table \ref{tab:glue} show that given the advantages in efficiency, the average performance of TLM is comparable to BERT across 8 tasks, which is consistent with our previous findings and demonstrates the effectiveness of TLM.



\section{Conclusions}

In this paper, we have proposed a simple, efficient, pretraining-free framework, \ours.
The core idea is to only use a tiny, task-relevant subset of the general corpus for language model training.
Our experiments show that \ours achieves results similar to or even better than PLMs, with a reduction of training FLOPs by two orders of magnitude. \ours opens the possibility of reducing the heavy reliance on large-scale PLMs and training a model from scratch in an efficient manner, while not hurting the overall performance.
We hope \ours will contribute to democratizing NLP and expediting its development by allowing most researchers to freely explore the architectures, loss functions, algorithms, and other design choices in the neighborhood of a state-of-the-art solution.

As discussed in Section \ref{sec:tlmplm}, there are several potential directions for future work.
It will be interesting to study how to use \ours to match the performance even larger-scale PLMs.
Moreover, further extending and improving \ours for few-shot and zero-shot learning is a crucial problem.


\bibliography{example_paper}
\bibliographystyle{icml2022}

\newpage
\appendix
\onecolumn

\counterwithin{figure}{section}
\counterwithin{table}{section}


\section{Comparison to Domain Adaptation}\label{sec:dapt_compare}

Our work is different from domain adaptation such as~\citet{dontstoppretraining}.
While domain adaptation aims to address how to effectively adapt a pretrained LM into one domain-specific task with sufficient domain data, this work targets to provide a method that is general enough to solve any task without domain data.
Nevertheless, we still compare \ours with~\cite{dontstoppretraining} as Table~\ref{tab:dapt} shows.
We hope to figure out that, under the harsh but practical condition that no domain data is accessible, whether our proposed framework \ours can still match or even outperform the traditional domain adaptation methods with large pretrained language models as well as domain data.

From results in Table~\ref{tab:dapt}, we have observations:
\begin{enumerate}
    \item We reproduced the RoBERTa-Base results using the hyper-parameters reported by~\citet{dontstoppretraining} as well as our own hyper-parameters. Results show that the baseline RoBERTa-Base results are underestimated in the paper with a gap of around 3 points. We list our hyper-parameters for fine-tuning RoBERTa in Table~\ref{tab:ft_hparams}.
    \item We also reproduced the DAPT+TAPT results using our own hyper-paraemters. Results show that DAPT+TAPT with new hyper-parameters also performs slightly better than it was reported by~\citet{dontstoppretraining}.
    \item From the perspective of total training computes (FLOPs), DAPT+TAPT consumes a comparable FLOPs with \ours(\textit{large-scale}), and \ours(\textit{large-scale}) achieved comparable results with DAPT+TAPT (i.e., 85.70 vs 85.57).
    However, from the perspective of data usage, DAPT+TAPT uses large amounts of domain data, the amount of which for each domain almost equals the amount of BERT total training corpus.
    \ours does not rely on it.
\end{enumerate}
\begin{table}[ht]
\caption{Comparison between the hyperparameters for fine-tuning from our implementation and from  \citet{dontstoppretraining}.}
\vspace{0.3cm}
    \begin{center}
    \centering
    \small
    \resizebox{0.5\linewidth}{!}{%
      \begin{tabular}{l|p{1.5cm}<\centering p{1.5cm}<\centering}
        \toprule[1pt]
        \tf{Hyper-parameters} & \tf{Ours} & \tf{Reported} \\
        \midrule[1pt]
        Epochs & - & 3 or 10 \\
        Training steps & 3e4 & - \\
        Patience & - & 3 \\
        Learning rate & 2e-5 & 2e-5 \\
        Batch size & 32 & 16 \\
        Max. grad. norm & - & 1 \\
        Weight decay & 0 & 0.1 \\
        \bottomrule[1pt]
        \end{tabular}
        }
      \end{center}
        \label{tab:ft_hparams}
\end{table}

\begin{table*}[htbp]
\caption{Comparison results of \ours and~\citet{dontstoppretraining}.}
    \begin{center}
    \centering
    \resizebox{\textwidth}{!}{%
    \begin{threeparttable}
      \begin{tabular}{l|ccccccccc}
        \toprule[1pt]
        & \tf{AGNews} & \tf{Hyp.} & \tf{Help.} & \tf{IMDB} & \tf{ACL.} & \tf{SciERC} & \tf{Chem.} & \tf{RCT} & \tf{Avg.} \\
        \midrule[1pt]

        RoBERTa-Base$^1$ & 93.90\std{0.20}&	86.60\std{0.90}&	65.10\std{3.40}&	95.00\std{0.20}&	63.00\std{5.80}&	77.30\std{1.90}&	81.90\std{1.00}&	87.20\std{0.10}&	81.25 \\
        RoBERTa-Base$^2$ & 93.97\std{0.13}&	88.50\std{4.18}&	67.45\std{0.49}&	95.43\std{0.07}&	63.87\std{1.24}&	79.97\std{1.29}&	81.50\std{0.94}&	87.26\std{0.08}&	82.24 \\
        RoBERTa-Base$^3$ & 94.02\std{0.15}&	93.53\std{1.61}&	70.45\std{0.24}&	95.43\std{0.16}&	68.34\std{7.27}&	81.35\std{0.63}&	82.60\std{0.53}&	87.23\std{0.09}&	84.12 \\
        
        \midrule
        DAPT$^1$& 93.90\std{0.20}&	88.20\std{5.90}&	66.50\std{1.40}&	95.40\std{0.10}&	75.40\std{2.50}&	80.80\std{1.50}&	84.20\std{0.20}&	87.60\std{0.10}&	84.00\\
        
        DAPT+TAPT$^1$& 94.60\std{0.10}&	90.00\std{6.60}&	68.70\std{1.80}&	95.60\std{0.10}&	75.60\std{3.80}&	81.30\std{1.80}&	84.40\std{0.40}&	87.80\std{0.10}&	84.75 \\
        
        DAPT+TAPT$^3$& 94.07\std{0.07}&	93.59\std{0.00}&	71.44\std{0.99}&	95.65\std{0.14}&	75.62\std{1.77}&	82.06\std{0.90}&	84.45\std{0.68}&	87.67\std{0.11}&	85.57 \\
        
        \midrule
        

        \makecell[l]{\ours \\ \xspace\textit{(large-scale)}}
        & 94.32\std0.07 
        & 95.16\std0.00 
        & 72.49\std0.33
        & 95.77\std0.24
        & 72.19\std1.72 
        & 83.29\std0.95
        & 85.12\std0.85 
        & 87.50\std0.12 & 85.74 \\

        \bottomrule[1pt]
        \end{tabular}
    \begin{tablenotes}
        \item[1] Results reported by \citet{dontstoppretraining}
        \item[2] Our reproduced results with the hyper-parameters reported by \citet{dontstoppretraining}
        \item[3] Results obtained by our own hyper-parameters
    \end{tablenotes}
\end{threeparttable}
}
\end{center}
\label{tab:dapt}
\end{table*}

\section{Detailed Experiment Settings}\label{sec:compuation}
\begin{table*}[htbp]
\caption{Detailed hyper-parameters for \ours of different scales for each task.}
    \begin{center}
    \centering
    \small
    \resizebox{\textwidth}{!}{%
    \begin{threeparttable}
      \begin{tabular}{l|l|cccccccc}
        \toprule[1pt]
        & \tf{Hyper-Parameters}& \tf{AGNews} & \tf{Hyp.} & \tf{Help.} & \tf{IMDB} & \tf{ACL.} & \tf{SciERC} & \tf{Chem.} & \tf{RCT} \\
        \midrule[1pt]
        \multirow{8}*{{\textit{\shortstack{Small\\ Scale}}}} &Top-$K$ & 50 & 5000 & 50& 500& 5000& 5000& 5000& 50\\
        &\eir & 1 & 99 & 1 & 19 & 999 & 999 & 999 & 3\\
        &\mw & 100 & 20 & 100 & 100 & 100 & 20 & 20 & 20\\
        &Source Corpus$^2$ 
        & \bertcorpus 
        & \bertcorpus
        & \bertcorpus 
        & \bertcorpus
        & \bertcorpus
        & \bertcorpus
        & \bertcorpus
        & \bertcorpus \\
        &Training Data Size$^3$ & 1.1GB & 0.2GB & 0.5GB& 0.9GB& 1.5GB& 1.6GB& 0.7GB& 0.8GB\\
        &Training Steps & 1E5 & 5E4 & 1.5E5 & 1.5E5 & 1.5E5 & 1.5E5 & 1.5E5  & 1E5\\
        &Batch Size & 256 & 256 & 256 & 256 & 256 & 256 & 256 & 256\\
        &Sequence Length& 128 & 128 & 128 & 128$^1$ & 128 & 128 & 128 & 128\\
        \midrule[1pt]
        \multirow{8}*{{\textit{\shortstack{Medium\\ Scale}}}} & Top-$K$ & 50 & 5000 & 50& 500& 5000& 5000& 5000& 50\\
        &\eir & 3 & 99 & 1 & 99 & 999 & 999 & 999 & 3\\
        &\mw & 100 & 100 & 1000 & 100 & 20 & 20 & 100 & 100\\
        &Source Corpus$^2$ 
        & \bertcorpus 
        & \bertcorpus
        & \bertcorpus
        & \bertcorpus
        & \bertcorpus
        & \bertcorpus
        & \bertcorpus
        & \bertcorpus \\
        &Training Data Size$^3$ & 1.1GB & 0.2GB & 0.5GB& 3.3GB& 1.5GB& 1.6GB& 0.7GB& 0.8GB\\
        &Training Steps & 3E5 & 1E5 & 3E5 & 3E5 & 3E5 & 3E5 & 3E5  & 3E5 \\
        &Batch Size & 256 & 256 & 256 & 256 & 256 & 256 & 256 & 256 \\
        &Sequence Length& 128 & 128 & 128 & 512 & 128 & 128 & 128 & 128 \\
        \midrule[1pt]
        \multirow{8}*{{\textit{\shortstack{Large \\ Scale}}}} & Top-$K$ & 100 & 10000 & 100& 1000& 10000& 10000& 10000& 100\\
        &\eir & 3 & 499 & 7 & 99 & 1999 & 1999 & 1999 & 7 \\
        &\mw & 100 & 20 & 100 & 1000 & 20 & 20 & 20 & 100\\
        &Source Corpus$^2$ 
        & \robertacorpus 
        & \robertacorpus
        & \robertacorpus
        & \robertacorpus
        & \robertacorpus
        & \robertacorpus
        & \robertacorpus
        & \robertacorpus \\
        &Training Data Size$^3$ & 3.1GB & 0.9GB & 1.7GB& 11GB& 3.5GB& 4.2GB& 2.5GB& 2.2GB\\
        &Training Steps & 5E5 & 3E5 & 5E5 & 5E5 & 5E5 & 3E5 & 5E5  & 5E5 \\
        &Batch Size & 256 & 512 & 512 & 512 & 512 & 512 & 256 & 256\\
        &Sequence Length& 128 & 128 & 128 & 512 & 128 & 128 & 128 & 128\\
        \bottomrule[1pt]
        \end{tabular}
\begin{tablenotes}
     \item[1] At a small scale on IMDB, we use a sequence length of 512 for internal data and a sequence length of 128 for external data.
     \item[2] \bertcorpus and \robertacorpus are our collected corpus that respectively match the original training corpus of BERT and RoBERTa.
     \item[3] \ours only uses a tiny subset of the source general corpus for training. We list the data size that are actually used for \ours training.
\end{tablenotes}
\end{threeparttable}
k,}
\end{center}
\label{tab:computation}
\end{table*}
Table~\ref{tab:computation} lists the detailed hyperparameters for TLM at stage 1 of different scales for each task. At small and medium scales, for tasks with less than 5K training examples (HyperPartisan, ChemProt, SciERC, ACL-ARC), we set $K = 5000$; for tasks with more than 100K training examples (RCT, AGNews, Helpfulness), we set $K = 50$, for the rest of the tasks (IMDB), we set $K = 500$. At the large scale, $K$ is doubled for each task.
At each scale on every task, we conduct grid search for $\rho_1 \in \{1,3,7,19,99,499,999,1999\}$ and $\rho_2 \in \{20, 100, 1000\}$, and adjust training steps, batch size and sequence length to minimize the training cost while preserving competitive performance. We observe that for almost all the tasks, the larger the training scale, the more reliance on external data, indicated by the increasing trend of $\rho_1$ and $\rho_2$ as the total training tokens goes up.

\section{Attention visualization on other tasks}
\label{sec:more_vis}
\begin{figure*}
\centering
\subfigure[\ours (\textit{Medium scale})]{
\begin{minipage}[t]{0.33\linewidth}
\centering
\includegraphics[width=2in]{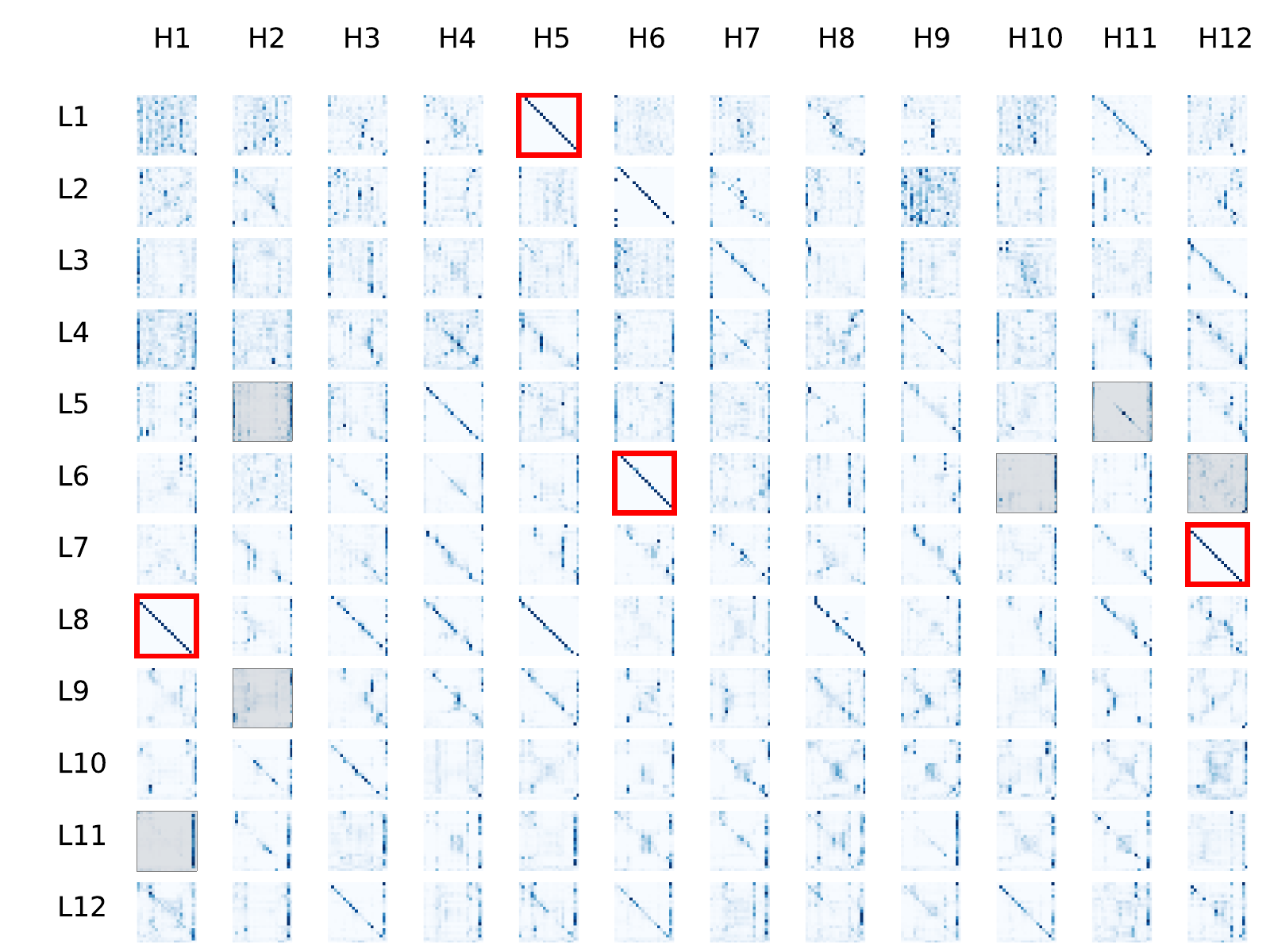}
\end{minipage}%
}%
\subfigure[BERT-Base]{
\begin{minipage}[t]{0.33\linewidth}
\centering
\includegraphics[width=2in]{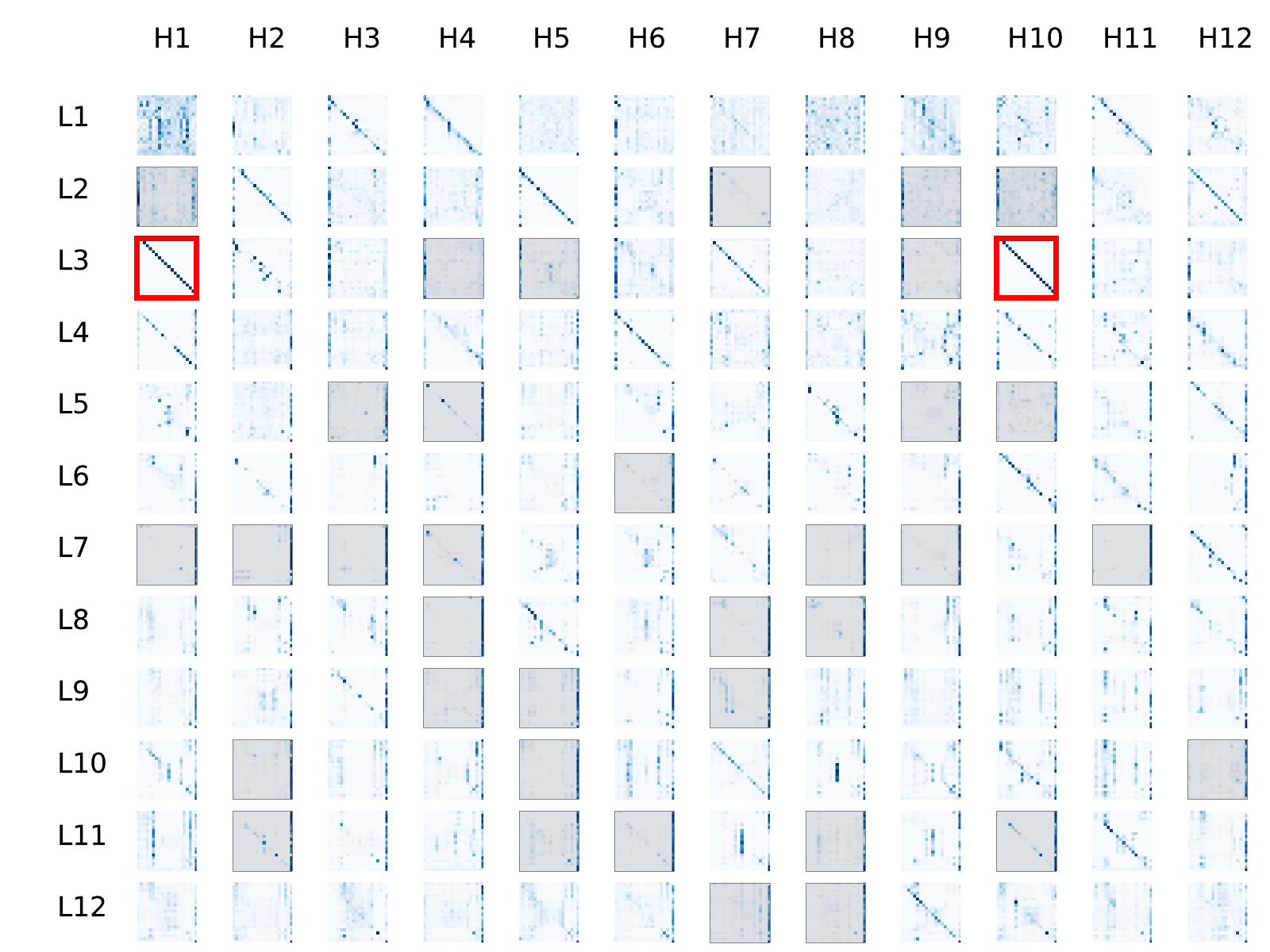}
\end{minipage}%
}%
\subfigure[RoBERTa-Base]{
\begin{minipage}[t]{0.33\linewidth}
\centering
\includegraphics[width=2in]{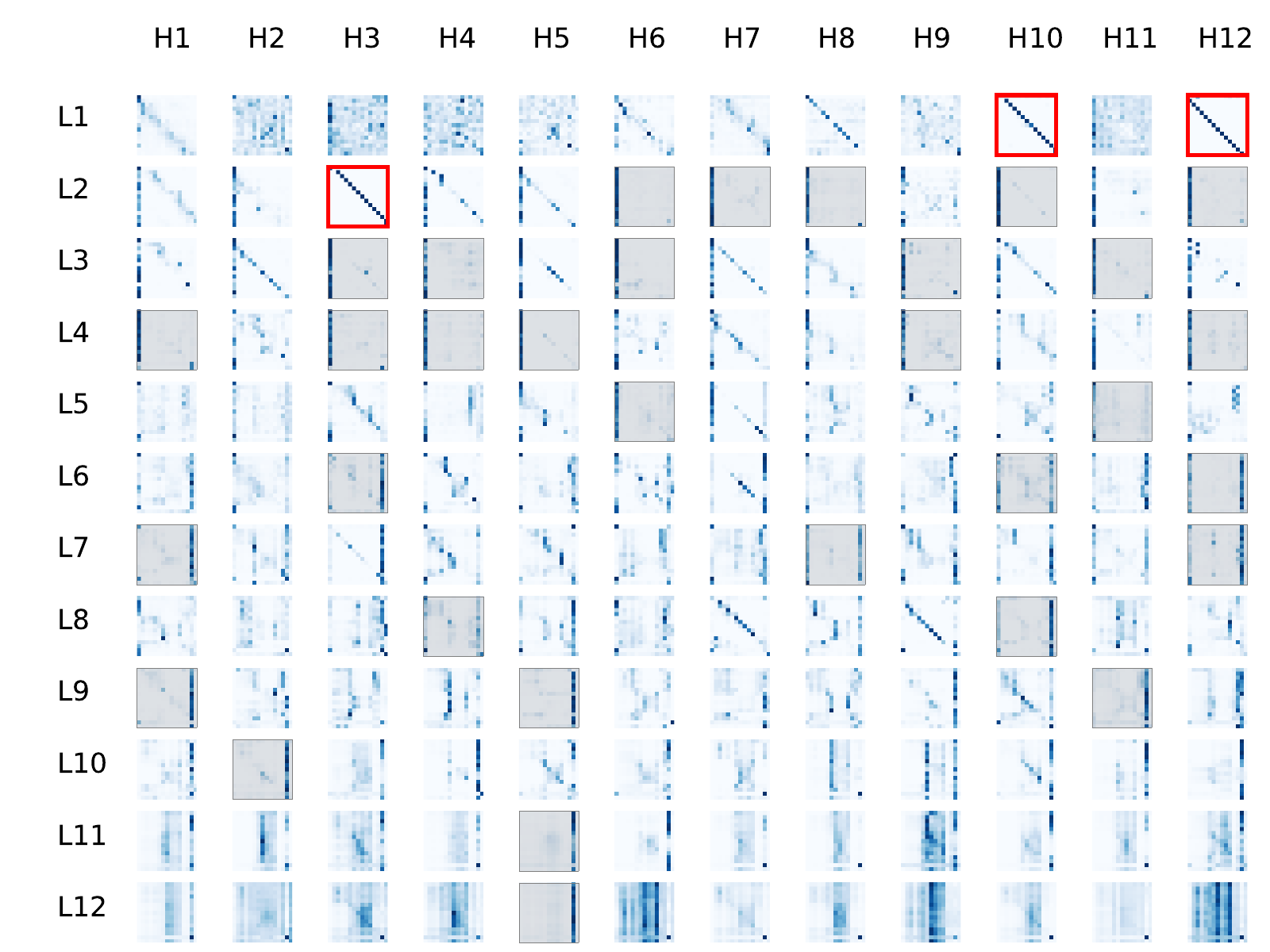}
\end{minipage}
}%
\caption{task: RCT ; input: "[CLS] twenty-eight individuals from outpatient physiotherapy departments were randomized. [SEP]"}
\label{RCT}
\end{figure*}

\begin{figure*}
\centering
\subfigure[\ours]{
\begin{minipage}[t]{0.33\linewidth}
\centering
\includegraphics[width=2in]{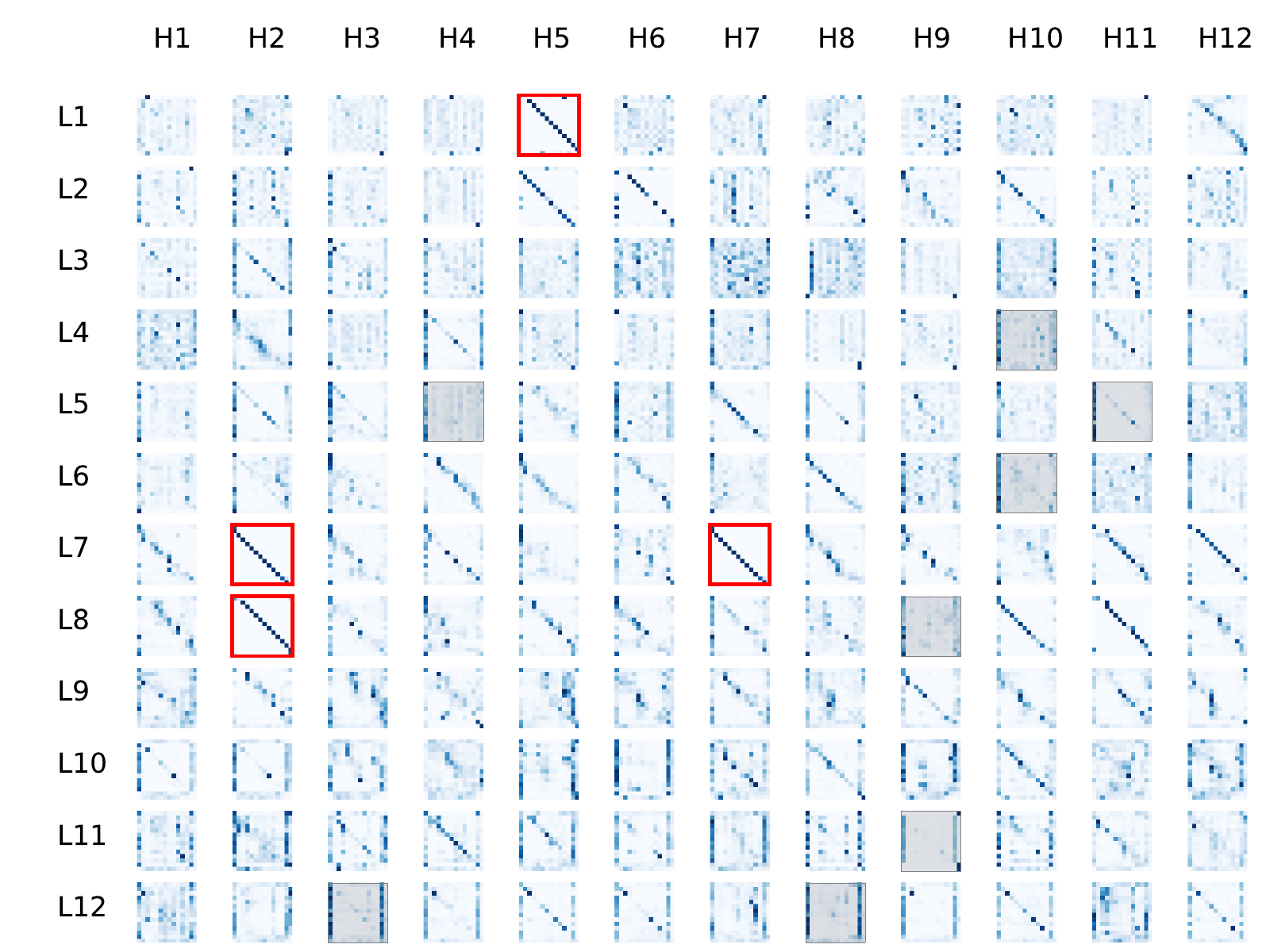}
\end{minipage}%
}%
\subfigure[BERT-Base]{
\begin{minipage}[t]{0.33\linewidth}
\centering
\includegraphics[width=2in]{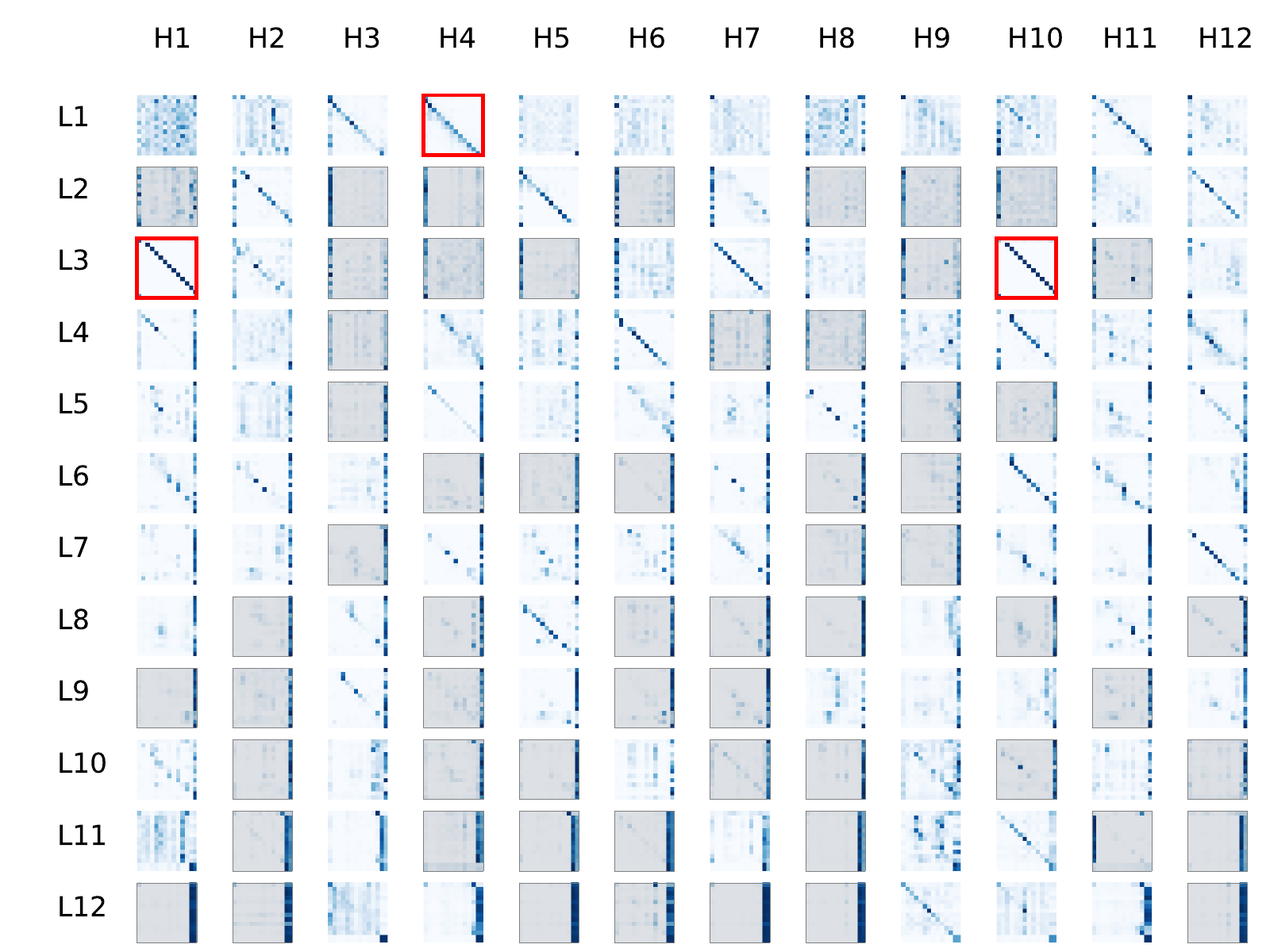}
\end{minipage}%
}%
\subfigure[RoBERTa-Base]{
\begin{minipage}[t]{0.33\linewidth}
\centering
\includegraphics[width=2in]{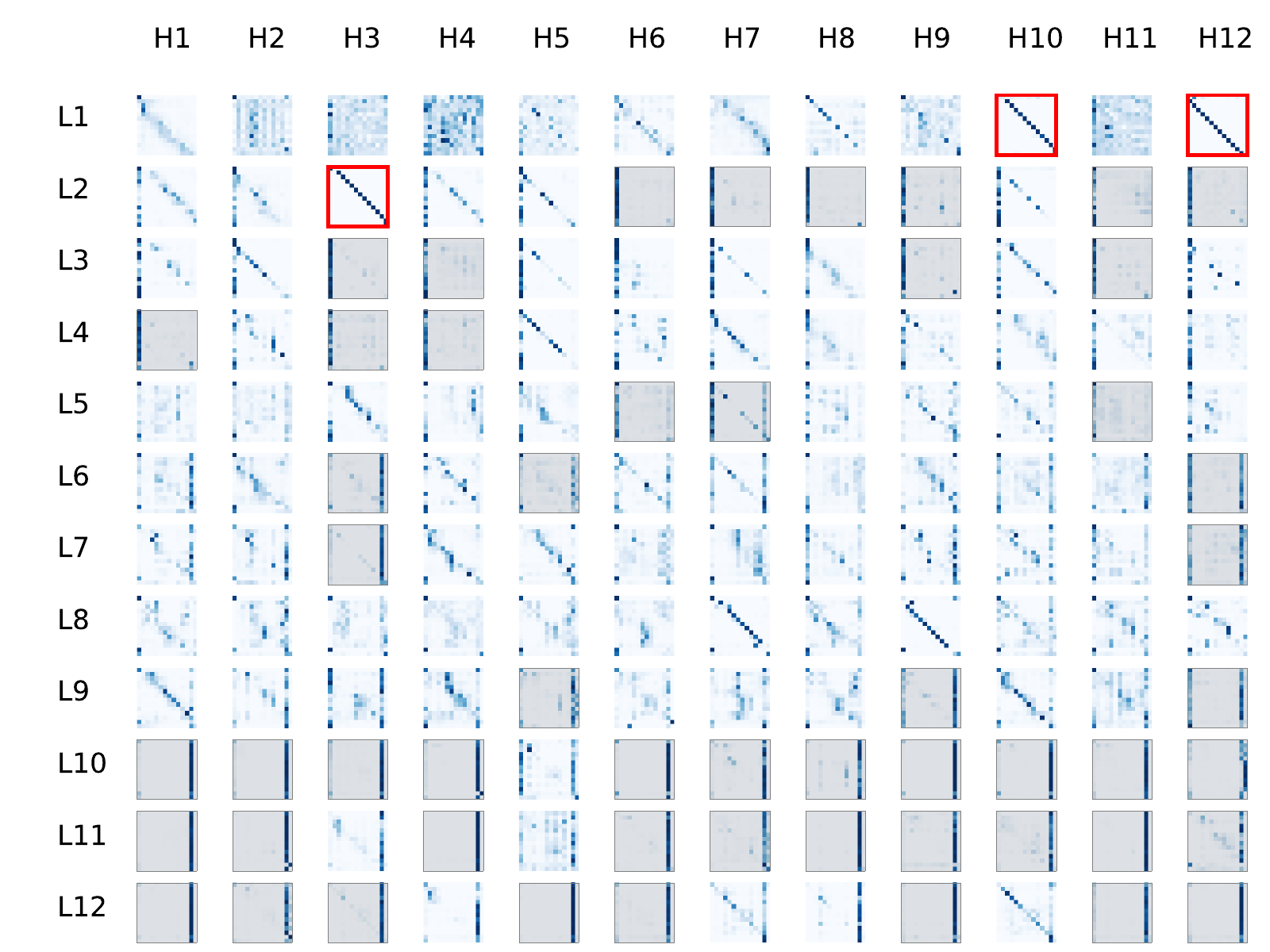}
\end{minipage}
}%
\caption{task: SciERC ; input: "[CLS] multi-view constraints associated with groups of patches are combined. [SEP]"}
\label{SciERC}
\end{figure*}

Besides ChemProt (Figure~\ref{visual}), we also experimented on RCT (Figure~\ref{RCT}) and  SciERC (Figure~\ref{SciERC}) to get attention visualizations. We find \ours consistently contains more positional heads (in red box) and less vertical heads (in gray mask). These results reveal that the aforementioned pattern generally holds for \ours. 

\end{document}